\def\year{2022}\relax
\documentclass[letterpaper]{article} 
\usepackage{aaai22}  
\usepackage{times}  
\usepackage{helvet}  
\usepackage{courier}  
\usepackage[hyphens]{url}  
\usepackage{graphicx} 
\usepackage{natbib}  
\usepackage{caption} 
\DeclareCaptionStyle{ruled}{labelfont=normalfont,labelsep=colon,strut=off} 
\usepackage{algorithm}
\usepackage{algorithmic}

\usepackage{newfloat}
\usepackage{listings}
\floatstyle{ruled}
\newfloat{listing}{tb}{lst}{}
\floatname{listing}{Listing}
\usepackage{latexsym}
\usepackage[T1]{fontenc}
\usepackage[utf8]{inputenc}
\usepackage{tablefootnote}
\usepackage{booktabs}       
\usepackage{amsfonts}       
\usepackage{nicefrac}       
\usepackage{microtype}      
\usepackage{lipsum}
\usepackage{graphicx}
\usepackage{color}
\usepackage{subfigure}
\usepackage{enumitem}
\usepackage{xcolor}
\usepackage{multirow}
\usepackage{array}
\usepackage{xspace}
\usepackage{dsfont}
\usepackage[switch]{lineno}

\newcommand{\hide}[1]{} 
\newcommand*{\eg}{\emph{e.g.},\@\xspace}
\newcommand*{\ie}{\emph{i.e.},\@\xspace}
\newcommand*{\etc}{\emph{etc.}\@\xspace}

\title{Drink Bleach or Do What Now? Covid-HeRA: A Study of Risk-Informed Health Decision Making in the Presence of COVID-19 Misinformation}

\author {
        Arkin Dharawat\textsuperscript{\rm 1},
        Ismini Lourentzou\textsuperscript{\rm 2},
        Alex Morales\textsuperscript{\rm 3},
        ChengXiang Zhai\textsuperscript{\rm 1} \\
}
\affiliations {
    \textsuperscript{\rm 1} Department of Computer Science, University of Illinois at Urbana - Champaign\\
    \textsuperscript{\rm 2} Department of Computer Science, Virginia Tech\\
    \textsuperscript{\rm 3} Twitter\\
    arkin.dharawat@gmail.com, ilourentzou@vt.edu, czhai@illinois.edu,
    alexmorales@twitter.com
}

\begin{document}
\maketitle
\begin{abstract}
Given the widespread dissemination of inaccurate medical advice related to the 2019 coronavirus pandemic (COVID-19), such as fake remedies, treatments and prevention suggestions, misinformation detection has emerged as an open problem of high importance and interest for the research community. Several works study health misinformation detection, yet little attention has been given to the perceived severity of misinformation posts. 
In this work, we frame health misinformation as a risk assessment task. More specifically, we study the severity of each misinformation story and how readers perceive this severity, \ie how harmful a message believed by the audience can be and what type of signals can be used to recognize potentially malicious fake news and detect refuted claims. To address our research questions, we introduce a new benchmark dataset, accompanied by detailed data analysis. We evaluate several traditional and state-of-the-art models and show there is a significant gap in performance when applying traditional misinformation classification models to this task. We conclude with open challenges and future directions.
\end{abstract}

\section{Introduction}\label{sec:intro}
While an increasing percentage of the population relies on social media platforms for news consumption, fake news and other types of misinformation have been also widely prevalent, putting audiences at great risk globally. Detecting and mitigating the impact of misinformation is therefore a crucial task that has attracted research interest, with a variety of approaches proposed, from linguistic indicators to deep learning models \cite{bal2020analysing}. 
Fake news frequently emerges for certain phenomena and topics, \eg public health issues, politics \etc \cite{allcott2019trends, shin2018diffusion, bode2018see}. Unsurprisingly, the same applies to the current global pandemic, where inaccurate stories are amplifying hate speech, increasing stigmatization, undermining public health response and polarizing public debate on topics related to COVID-19. It is often difficult for users, who may decide to take action based on health advice found online, to understand the consequences and potential risks from following unreliable guidance, especially when all information spread by influential users is perceived as equally credible \cite{morales2020crowdqm}. This adds to the worry and anxiety felt by many, already in a difficult situation \cite{kleinberg2020measuring}. 

Moreover, reducing the impact of misinformation on potential health-related decisions requires deeper understanding of the target audience, which goes beyond fact-checking. To this end, we aim to capture how social media users view the severity of misinformation circulating online. Although much work has been focused on identifying health-related misinformation, there has been little attention to further making a distinction between the perceived severity of misinformation \cite{fernandez2018online}. 
Severity varies greatly across each message: some might be jokes, some might be discussing the impact of fake news or refute the claim, others might be highly malicious, while others might be simply inaccurate information with limited effects. Overall, the severity of each message can vary depending on its content, \eg urging users to eat garlic is less severe than urging users to drink bleach which has led to hundreds of deaths \cite{islam2020covid}.

Some news articles related to COVID-19 misinformation reporting hospitalization, injuries, and deaths may be interpreted as weak signals of misinformation severity to some extent.
However, how readers view the severity level for each misinformation story remains a challenging task that has not been previously studied, even though building such models would be beneficial, \eg in prioritizing policy-making towards mitigating severely harmful misinformation, such as potentially deadly fake remedies; a goal that we hope to achieve in this paper.

Furthermore, to mitigate misinformation spread, it is critical to build systems that inform the target audience about the negative impacts and provide resources for properly assessing information early in the process. 
Creating an early warning system for misinformation requires detection mechanisms directly based on content. Propagation-based methods that rely on social context information by incorporating additional signals such as like, retweets and network information \cite{zhou2020fake} are often inadequate for early detection. In addition, adversarial attacks of misinformation spreaders due to manipulation strategies for both networks and content, \eg forming groups and increasing influence by connecting with each other, undermine the performance of existing approaches \cite{wu2019misinformation}.

Based on these challenges, our work aims to help reduce the impact of health misinformation and enable better decision-support health information systems, by proposing a novel perceived health risk misinformation framework. Specifically, we aim at understanding the impact of unreliable advice from inaccurate sources based on the perceived severity of misinformation. To this end, we also introduce a new benchmark dataset that contains social media posts annotated on a Likert scale, based on whether the  content of the post is perceived to be: a) Real News/Claims, b) Refutes/Rebuts, c) Misinformation (Other), d) Misinformation (Possibly severe), and e) Misinformation (Highly severe). The distinction of different reader-perceived severity levels is judged based on their potential to impact a user's health, assuming the individual might be making decisions upon reading the advice and/or suggestions in the post. In other words, our goal is to produce a label that reflects the level of recognized risk factors in the presence of inaccurate claims and news, on a challenging cold-start content-based scenario where additional metadata information such as likes, retweets and user information is not available, and conditioned on the worst-case assumption that the user will follow the advice. We conduct the first empirical study of the effectiveness of representative machine learning methods for predicting the perceived severity of misinformation so as to establish baseline benchmark results for facilitating further research. 
 
Our contributions can be summarized as follows:
\begin{itemize}
    \item In contrast with prior work that treats misinformation as a fact-checking and/or veracity task, we introduce a new task of modeling the perceived risk of health-related misinformation.
    
    \item  To facilitate research on this new task and understand how users view harmful unreliable information shared online, we release a novel social media \textbf{Covid} \textbf{He}alth \textbf{R}isk \textbf{A}ssessment (\textbf{\texttt{Covid-HeRA}}) misinformation dataset that consists of $61,286$ tweets labeled on a severity scale.

    \item We benchmark several content-based classification models, including state-of-the-art fake news detection models and contextualized embeddings.  Experimental results show a significant performance gap when compared to traditional misinformation detection settings. 
 
    \item We present a detailed data analysis that reveals several key insights about the most prominent unreliable news. Moreover, we present technical and modeling challenges that this new task poses.
\end{itemize}

Our dataset captures how social media users perceive severe misinformation and can potentially be used to determine the user's understanding and recognition of the severity of misinformation.
We hope that \textbf{\texttt{Covid-HeRA}} will facilitate future research on developing detection models that take into account user views and misinformation severity\footnote{Covid-HeRA is open-sourced at {\url{https://github.com/TIMAN-group/covid19_misinformation}}}.

\section{Related Work}\label{sec:rel}
Health-related misinformation research spans a broad range of disciplines including computer science, social science, journalism, psychology, and so on \cite{dhoju2019differences,castelo2019topic,fard2020misinformation}. While health-related misinformation is only a facet of misinformation research, there has been much work analysing misinformation in different medical domains, such as cancer \cite{bal2020analysing, loeb2019dissemination}, orthodontics \cite{kilincc2019assessment}, sexually transmitted diseases and infections \cite{zimet2013beliefs,lohmann2018saying, joshi2018shot,tomaszewski2021identifying}, autism \cite{baumer2019speaking}, influenza \cite{culotta2010towards,signorini2011use}, and more recently COVID-19 \cite{garrett2020covid, brennen2020types, cinelli2020covid, cui2020coaid}. 
Social media data have also been used to monitor influenza prevalence and awareness \cite{smith2016towards, ji2013monitoring, huang2017examining}.
Systems such as Google Flu Trends use real-time signals, such as search queries, to detect influenza epidemics \cite{ginsberg2009detecting, preis2014adaptive, santillana2014can,kandula2019reappraising}.
However, relying solely on search queries leads to an overestimation of influenza, namely because there is no distinction between general awareness about the flu and searches for treatment methods \cite{smith2016towards,klembczyk2016google}.
Our work focuses on social media, in particular health misinformation on micro-blogging sites, such as Twitter. 

\citet{tomeny2017geographic} examined geographic and demographic trends in anti-vaccine tweets. \citet{huang2017examining} examine the geographic and demographic patterns of the flu vaccine in social media. Recent research has also focused on identifying users disseminating misinformation, in the case of cancer treatments \cite{ghenai2018fake}, as well as hybrid approaches combining user-related features with content features \cite{ruchansky2017csi}. 
\citet{joshi2018shot} make a distinction between \textit{vaccine hesitancy} identification, and \textit{vaccination behavior detection}, in that the former deals with the attitudes, or stance, while the latter is concerned with detecting the action of getting vaccinated. Our work models how users perceive different severity levels of misinformation.

With the threat of COVID-19 misinformation to public health organizations, there have been several call-to-actions \cite{chung2020ct, mian2020coronavirus, calisher2020statement} to underscore the gravity and impact of COVID-19 misinformation \citet{garrett2020covid}. \citet{tasnimimpact} outlines several potential strategies to ensure effective communication on COVID-19, \eg ensuring up-to-date reliable information via identifying fake news and misinformation.
\citet{brennen2020types} analyze the different types, sources, and claims of COVID-19 misinformation, and show that the majority appear on social media outlets. As the dialog on the pandemic evolves, so does the need for reliable and trustworthy information online \cite{cuan2020misinformation}.

\citet{pennycook2020fighting} show that people tend to believe false claims about COVID-19 and share false information when they do not think critically about the accuracy and veracity of the information. Recent work aims at understanding reader perception of news reliability \cite{gabriel2021misinfo}. \citet{kouzy2020coronavirus} show that $\sim25\%$ of COVID-19 messages contain some form of misinformation and $\sim17\%$ contain some unverifiable information. \citet{singh2020first} provide a large-scale exploratory analysis of how myths and COVID-19 themes are propagated on Twitter, by analyzing how users share URL links. 
\citet{cinelli2020covid} cluster word embeddings to identify topics and measure the engagement of users on several social media platforms. They provide a comparative study of information reproduction and provide rumor amplification parameters for COVID-19 on these platforms.

\begin{table*}[th!]
\centering
\resizebox{0.95\textwidth}{!}{%
\fontsize{10}{10}
    \begin{tabular}{p{0.16\linewidth}p{0.53\linewidth}p{0.4\linewidth}}
    \toprule
    \textbf{Category} & \textbf{Tweet} & \textbf{Reasoning} \\
    \midrule
    \textbf{Real News/Claims} & ``\textit{What is a coronavirus? Large family of viruses. Some can cause illness in people or animals. In humans, it's known to cause respiratory infections}'' &  This category includes correct facts and accurate news. \\
    \midrule
    \textbf{Possibly severe} & ``\textit{Vitamin C Protects Against Coronavirus}'' &  Although an individual may decide to take daily doses of vitamin C, it is unlikely to be harmful and potential risks are less significant than for other actionable items. \\ 
    \midrule
    \multirow{3}*{\textbf{Highly severe}} & ``\textit{Good News: Coronavirus Destroyed By Chlorine Dioxide \_ Kerri Rivera}''  & \multirow{3}{\linewidth}{These tweets either promote specific behavioral changes and fake remedies with increased health risks, or may result in increased exposure for certain socioeconomic groups.} \\
    & ``\textit{Flu Vaccine Increases Coronavirus Risk 36\% Says Military Study}'' & \\
    &  ``\textit{People of color may be immune to the Coronavirus because of Melanin blackmentravels}'' & \\
    \midrule
    \textbf{Other} & ``\textit{Vatican confirms Pope Francis and two aides test positive for Coronavirus}'' &  This involves other types of misinformation not related to health, \eg conspiracy theories and public figure rumors; health-related behavioral changes might be less likely to occur.\\
    \midrule
    \textbf{Refutes/Rebuts Misinf/tion} & ``\textit{For those who think COVID-19 is just like the flu: In the 2018-19 flu season, there were 34,000 deaths over a period of months and months. In contrast, there have been over 40,000 COVID-19 deaths in a little more than a month and half even with social distancing.}'' & These type of posts are useful in identifying fake claims, as well as presenting opposing views.\\
  \bottomrule
\end{tabular}%
}
    \caption{COVID-19 example tweets labeled on a severity scale, alongside with a brief explanation of each category.}
    \label{tab:examples}
\end{table*}

The coronavirus pandemic has led to several measures enforced across the globe, from social distancing and shelter-in-place orders to budget cuts and travel bans \cite{nicola2020socio}. News circulates daily advice for the public. Some articles, however, contain fake remedies that reportedly cure or prevent COVID-19, promote false diagnostic procedures, report incorrect news about the virus properties or urge the audience to avoid specific food or treatments that might supposedly either make symptoms worse or the reader more likely to contract the virus\footnote{\url{https://www.cmu.edu/ideas-social-cybersecurity/research/coronavirus.html}}. With such information overload, any decision-making procedures based on misinformation have a high likelihood of severely impacting one's health \cite{ingraham2020fact}. 
Therefore, we aim to capture how users perceive the severity of incorrect information released on social media, as well as to detect any posts that refute or rebut coronavirus-related false claims. To this end, we release a novel dataset, present experimental results with several state-of-the-art machine learning models and conclude with possible future extensions.
\section{Methodology}\label{sec:meth} 
In this section, we introduce our task formulation, data collection \& annotation strategy. Subsequently, we present detailed statistics and analysis with key insights on the most prevalent harmful misinformation online.
We frame the task as multi-class classification and design an annotation scale, based on whether a social media post has the potential to guide the audience towards health-related decisions or behavioral changes with high-risk factors, \ie high likelihood of severely impacting one's health. In other words, we aim to model the \textit{perceived severity} by ordinary users that would potentially follow medical advice found online. Since, to the best of our knowledge, there is no public dataset appropriate for this task, we release \texttt{Covid-HeRA} to facilitate future research.

Upon extensive team discussions, and in  conjunction with data inspection, we design a five-class framework that is inspired by Likert-type scales \cite{likert1932technique}, typically used in risk stratification and ordinal risk rating. To this end, each social media post is categorized as follows: a) \textbf{Real News/Claims}, \ie messages with reliable correct information, b)  \textbf{Refutes/Rebuts}, \ie messages which contain a refutation or rebuttal of an incorrect statement, c) \textbf{Other}, \ie other types of misinformation or misinformation that is unlikely to result in risky behavioral changes, d) \textbf{Possibly severe} misinformation, with possible severe health-related impact and e) \textbf{Highly severe} misinformation with increased potential risks for any individual following the advice and suggestions expressed in the social media post content. These categories enable researchers to study the impact of coronavirus health misinformation at a finer granular level, to develop algorithms that caution the audience on the potential risks and to design systems that present unbiased information, \ie both the original - potentially unreliable - post, along with any possible rebutting claims expressed online. In Table \ref{tab:examples}, we present example posts for each category.

We make use of CoAID~\cite{cui2020coaid}, a large-scale healthcare misinformation data collection related to COVID-19, with binary ground-truth labels for news articles and claims, accompanied with associated tweets and user replies. This dataset provides us with a large amount of reliable Twitter data and alleviates the need for labeling tweets as real or fake. Furthermore, it has the potential to be updated automatically with additional instances, enabling semi-supervised models as future work. 
To obtain annotations based on our defined severity categorization, all CoAID tweets labeled as misinformation are shuffled and distributed to two different annotators. Our annotators are ordinary social media users, since we want to capture how regular users perceive potentially harmful misinformation shared online. This formulation is useful in studying target audience behavioral aspects influenced by misinformation messages and in rethinking platform policies for managing misinformation based on perceived severity levels.

\begin{figure}[t!]
\centering
\subfigure{
\includegraphics[width=0.5\columnwidth]{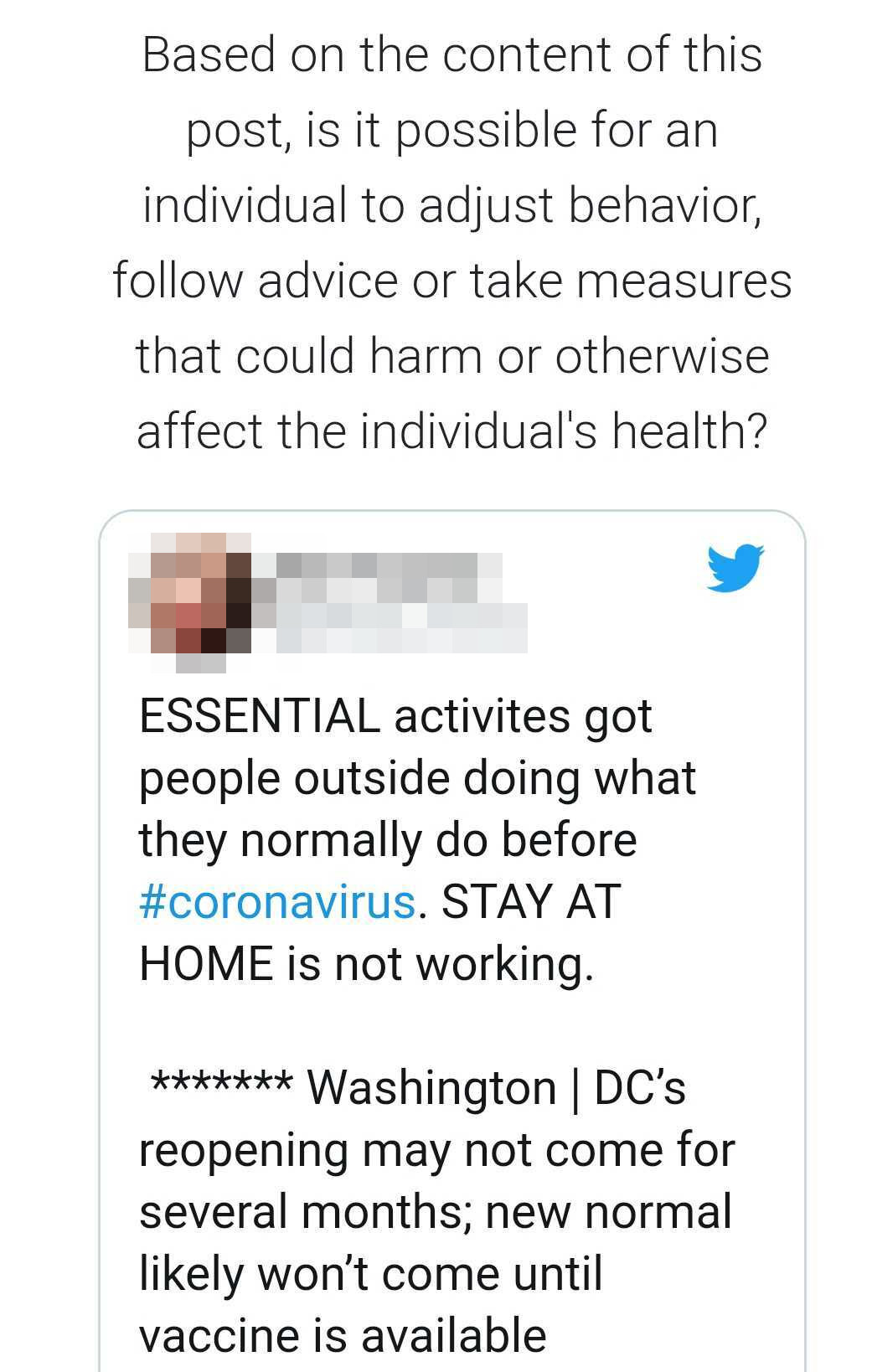}
\includegraphics[width=0.5\columnwidth]{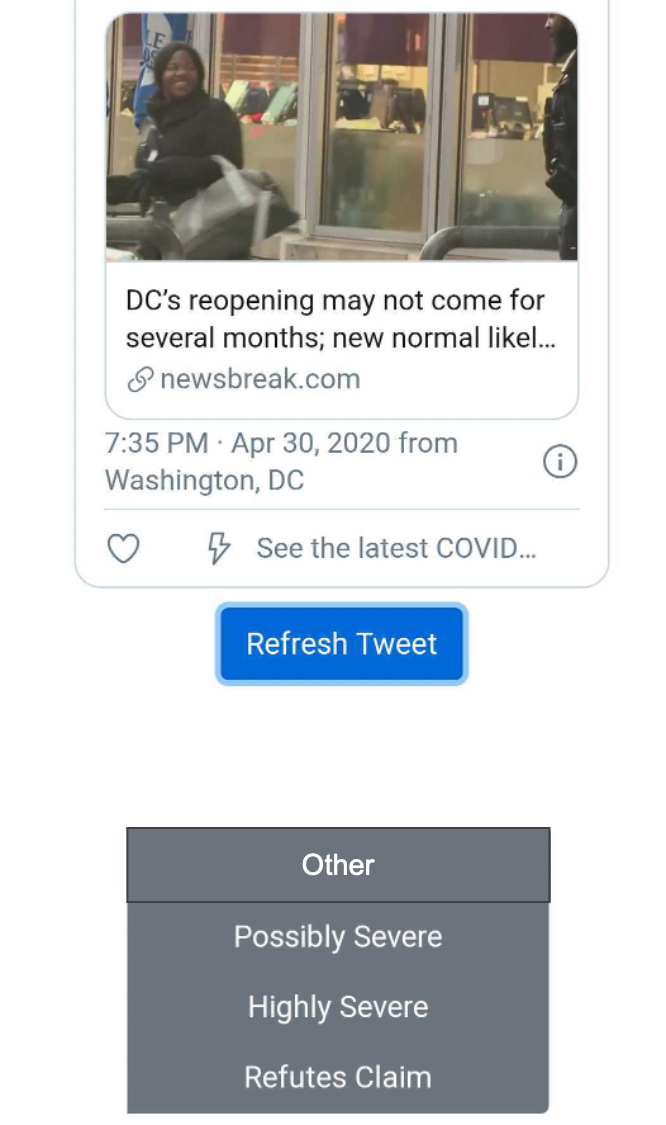}
}
\caption{A screenshot of a tweet example and possible annotation options.} 
\label{fig:datacollection}
\end{figure}

Each annotator is asked to judge whether any decisions can be taken or other actionable items can be performed based on the expressed content and whether those could result in harmful choices, risky behavioral changes or other severe health impacts. Additionally, we asked annotators to flag any post that expresses an opinion or argument against the unreliable claims, \ie refutes or rebuts misinformation (see Figure \ref{fig:datacollection} for a screenshot of our annotation interface). Each tweet is annotated jointly by the two annotators. To resolve conflicts on ambiguous instances, a final round of annotator discussions was introduced as an additional step, where an external validator provided an additional label after the discussion. To assess agreement levels, the external validator was asked to also annotate a random sample of the labeled tweets. Cohen's kappa coefficient between the annotators and the validator was $0.7037$, which shows good agreement on the task \cite{hunt1986percent}.  Both annotators and the validator are proficient in English, and  diversified w.r.t. nationality to  mitigate systemic bias as much as possible.

The total number of tweets labeled per category, alongside the number of unique words, are presented in Table \ref{tab:dataset}. In addition, the dataset contains a \textit{claim index}, with each tweet mapped to a specific category claim that captures the main topic\footnote{Such mapping is already provided by the CoAID dataset.}. The average time between a tweet and a reply is approximately 11hrs for ``Real News/Claims'' ($40,333$ seconds), 12hrs for ``Other'' ($44,708$ seconds), 8hrs for ``Possibly severe'', 9hrs for ``Highly severe'' ($32,369$ seconds) and 7hrs for ``Refutes/Rebuts'' ($25,238$ seconds). 

\begin{table}[t!]
\centering
\resizebox{0.9\columnwidth}{!}{%
\begin{tabular}{ccc}
\toprule
Category & \#Tweets & \#Tokens (Vocab) \\ \midrule
\textbf{Possibly severe} & 439 & 11,171 \\
\textbf{Highly severe} & 568 & 16,328 \\
\textbf{Refutes/Rebuts} & 447 & 2,244 \\
\textbf{Other} & 1,851 & 3,324 \\
\textbf{Real News/Claims}  & 57,981 & 51,478 \\
\midrule 
\midrule 
\textbf{Total} & 61,286 & 84,545  \\\bottomrule
\end{tabular}
}
\caption{\texttt{Covid-HeRA} Dataset Statistics.}
\label{tab:dataset}
\end{table}

\subsection{Data Analysis}\label{sec:dataanalysis} 
\begin{figure*}[th!]
\centering
\subfigure[Possibly severe]{
\includegraphics[width=0.39\columnwidth]{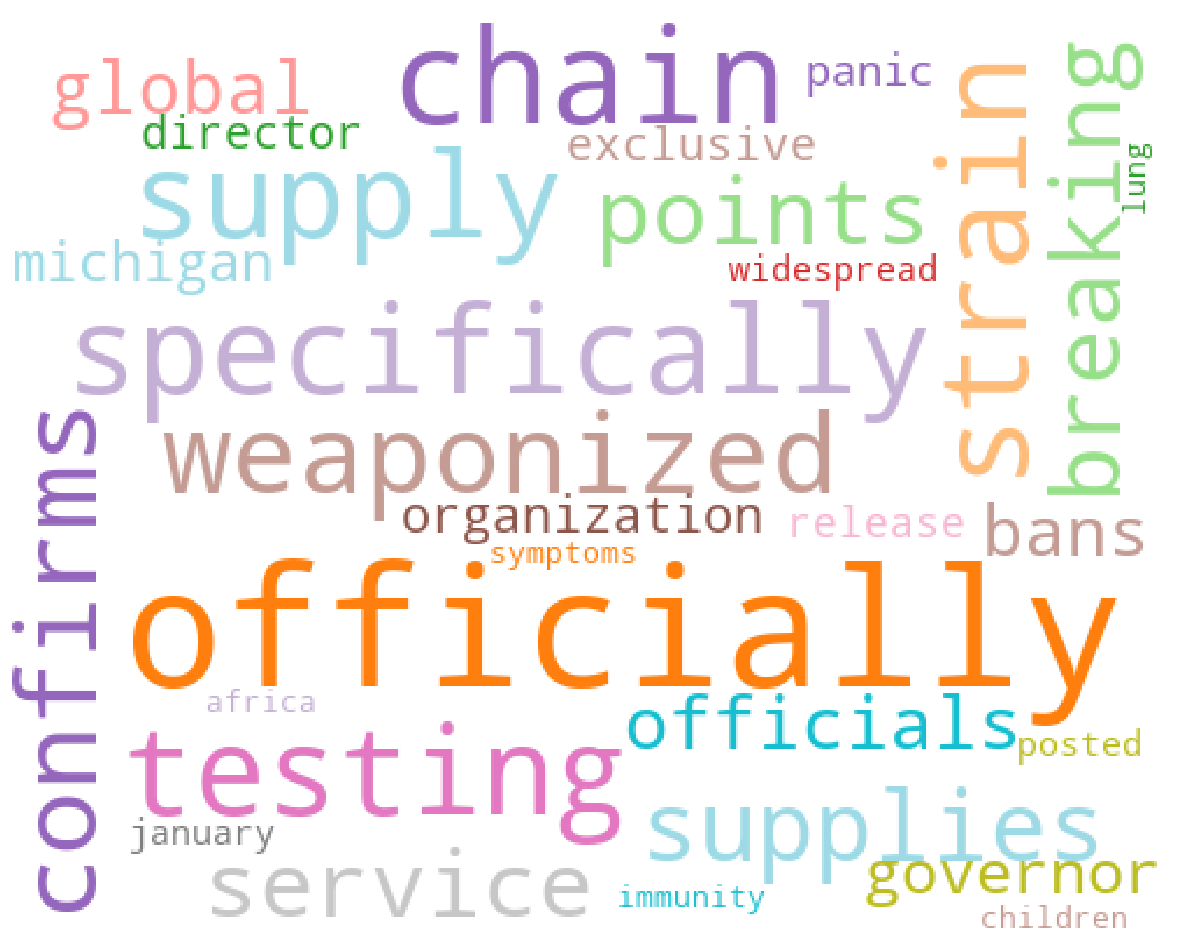}}
\subfigure[Highly severe]{
\includegraphics[width=0.39\columnwidth]{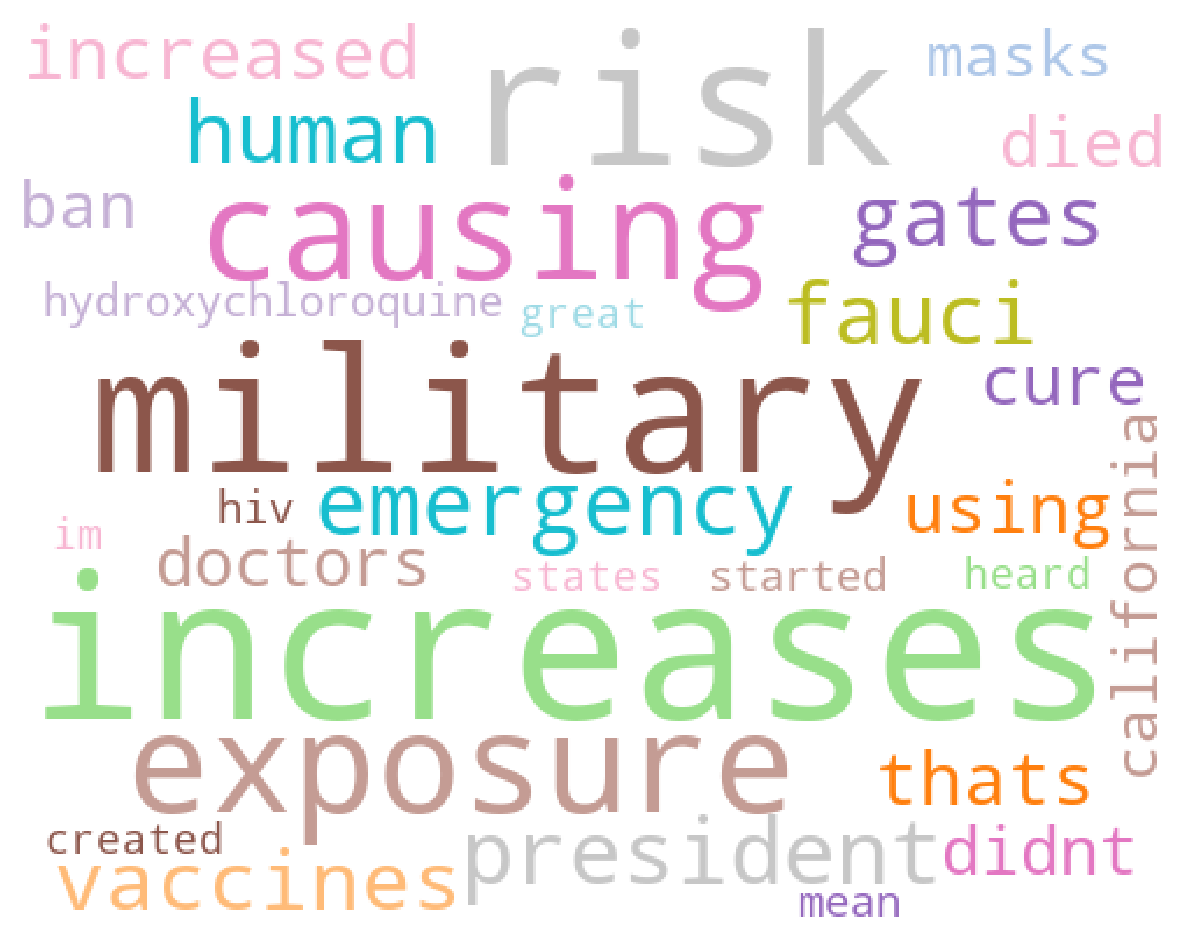}}
\subfigure[Refutes/Rebuts]{
\includegraphics[width=0.39\columnwidth]{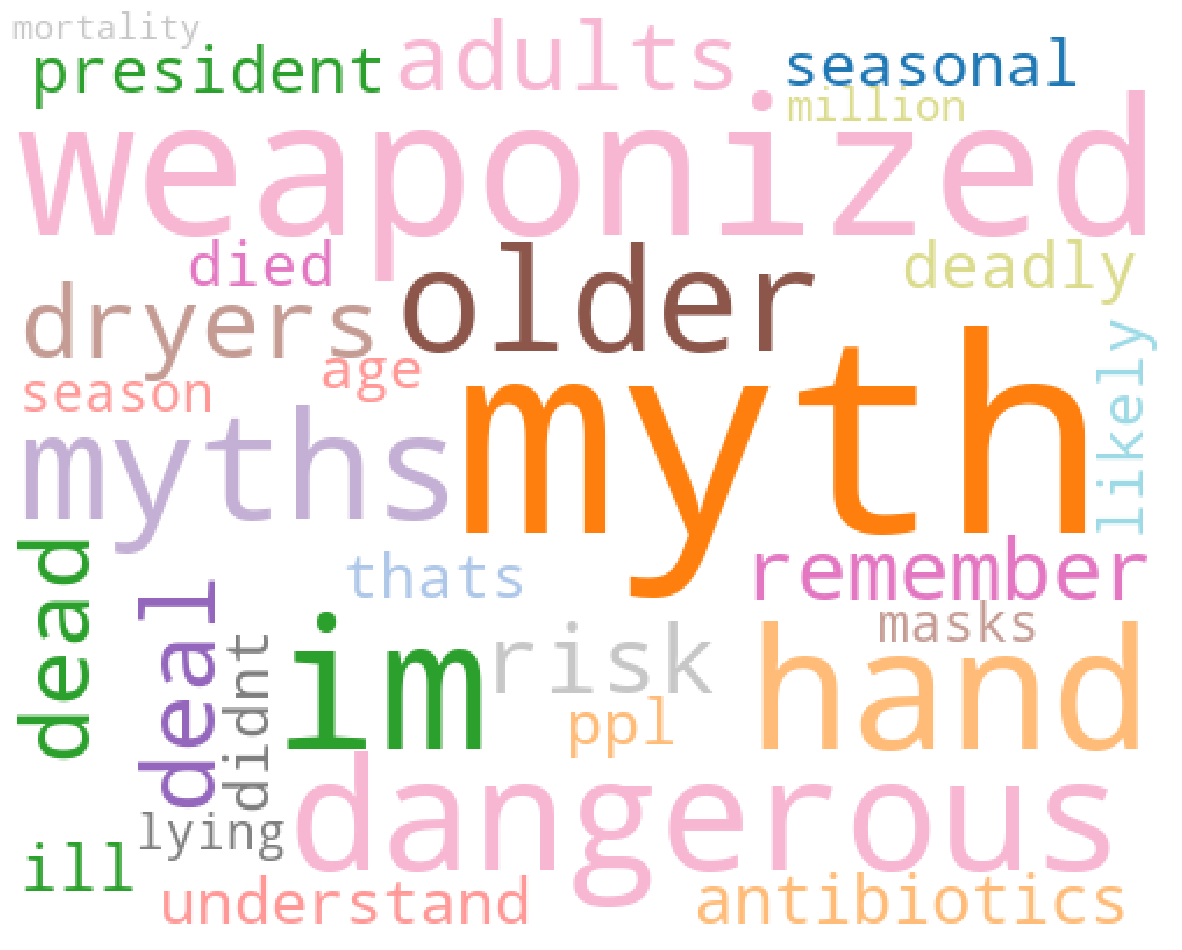}}
\subfigure[Other]{
\includegraphics[width=0.39\columnwidth]{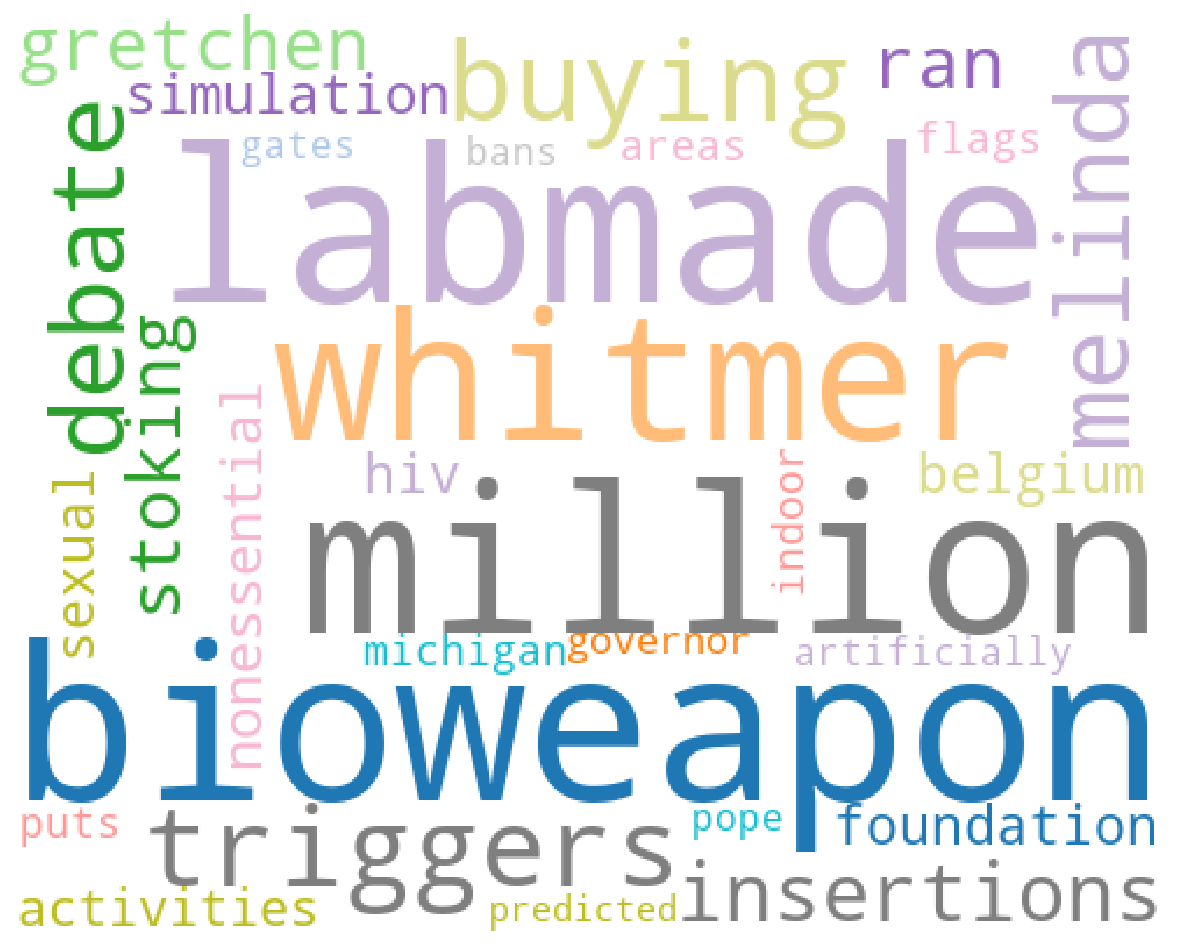}}
\subfigure[Real News/Claims]{
\includegraphics[width=0.39\columnwidth]{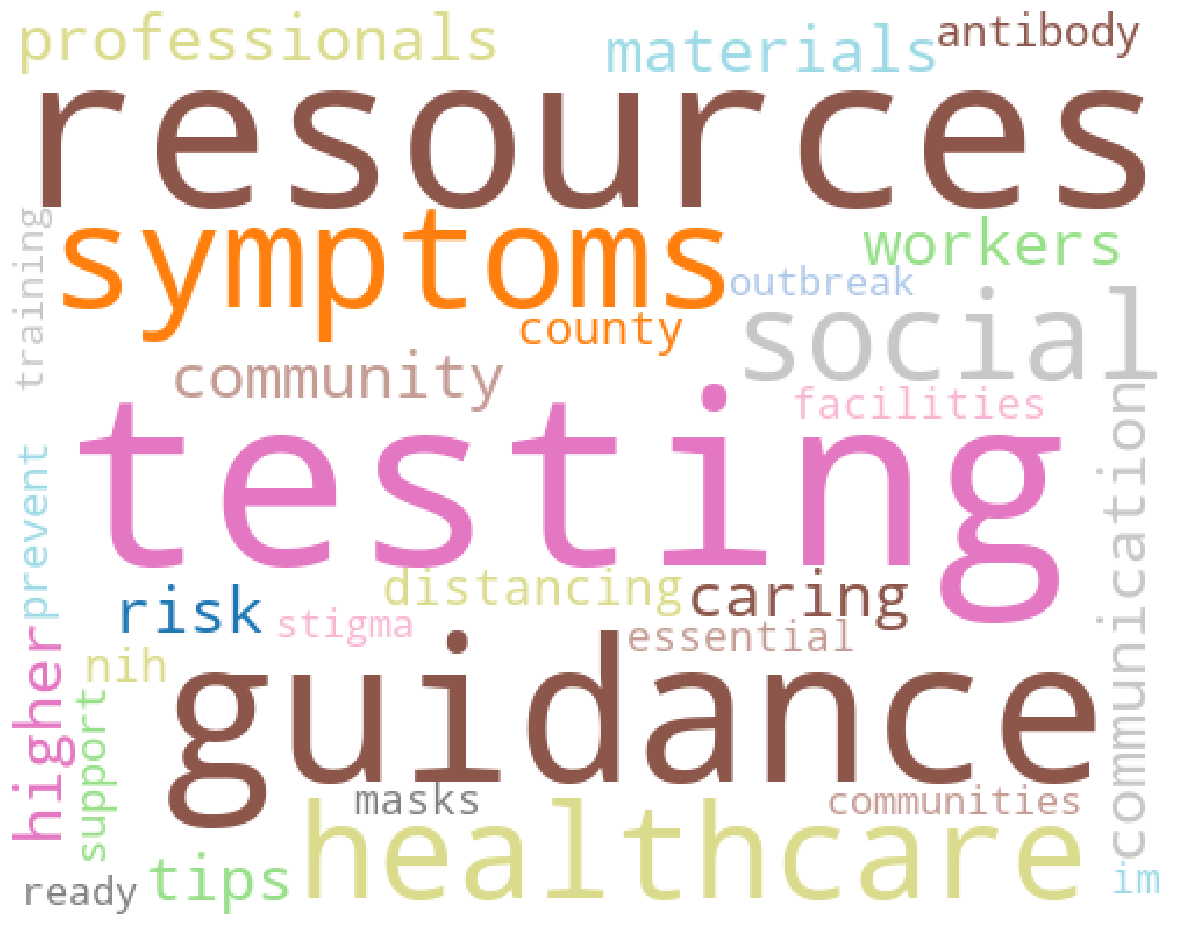}}
\caption{Most common terms per category}  
\label{fig:freq}
\end{figure*}

We first identify the most frequent discriminative terms per category, \ie terms that appear very often in a specific category, but may be infrequent in the remaining categories.  We use a  $0.5\%$ document frequency threshold to discard terms that are very common across the whole data collection, \eg ``\textit{COVID-19}'' or ``\textit{virus}'' appearing in more than half of the tweets. In Figure \ref{fig:freq}, we visualize the top-30 terms per category, with each term weighted by its representativeness. When comparing the ``Other'' category with the rest of the categories, we see that many of the terms refer to conspiracy theories about COVID-19, such as ``\textit{artificially}'', ``\textit{labmade}'', ``\textit{bioweapon}'' which pertains to the conspiracy that COVID-19 is a man-made virus. The top terms for the ``Highly severe'' category seem to be about treatments and are more risky words such as ``\textit{risk}'', ``\textit{mask}'', ``\textit{cure}'', ``\textit{vaccines}'', and ``\textit{hydroxychloroquine}''. The top terms for the “Refutes/Rebuts” contain keywords such as ``\textit{myths}'', ``\textit{weaponized}'', ``\textit{lying}'', and ``\textit{antibiotics}'' as the messages in this category address and debunk conspiracies and misinformation, while the top terms for the ``True News/Claims'' are ``\textit{resources}'', ``\textit{symptoms}'', ``\textit{testing}'', and ``\textit{guidance}'', as these messages are generally informative and provide advice about COVID-19. 

\begin{figure}[t!]
\centering
\includegraphics[width=0.6\columnwidth]{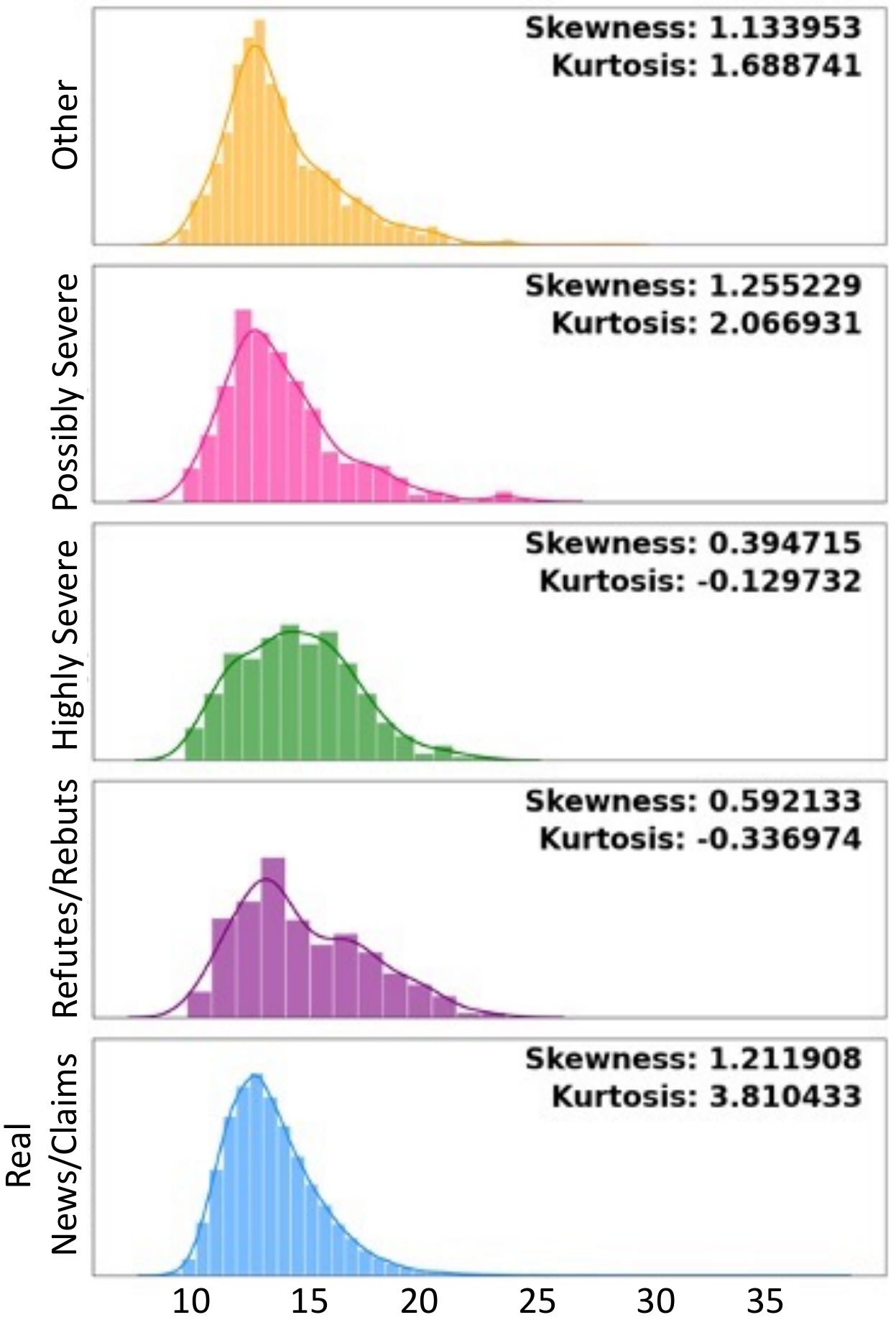}
\caption{Category compactness, measured by the distance distribution of Twitter posts, relative to its category centroids.} 
\label{fig:compact}
\end{figure}

In Figure \ref{fig:compact}, we visualize the compactness of each category. We use pre-trained BERT embeddings \cite{devlin2018bert} to measure how close each tweet is by the centroid of its corresponding category. More specifically, we compute the centroid as the arithmetic mean of all tweet embeddings from a specific category.
We hypothesize that compact categories are more likely to be well-formed and thus easier to classify. We compute the skewness and kurtosis for each distribution. Each distribution shows a positive skewness, and as we can see, they are right-tailed distributions. The ``Possibly severe'' category is the most skewed and the ``Highly severe'' category is the least. The negative kurtosis of both the ``Highly severe'' and ``Refutes/Rebuts'' categories shows that these categories have less of a peak and appear more uniform, which is also evident by the flatness of these curves. This may be due to the broad range of topics covered in both of these categories compared to the rest. 

\begin{figure*}[t!]
\centering
\subfigure[Possibly severe]{
\includegraphics[width=0.4\columnwidth]{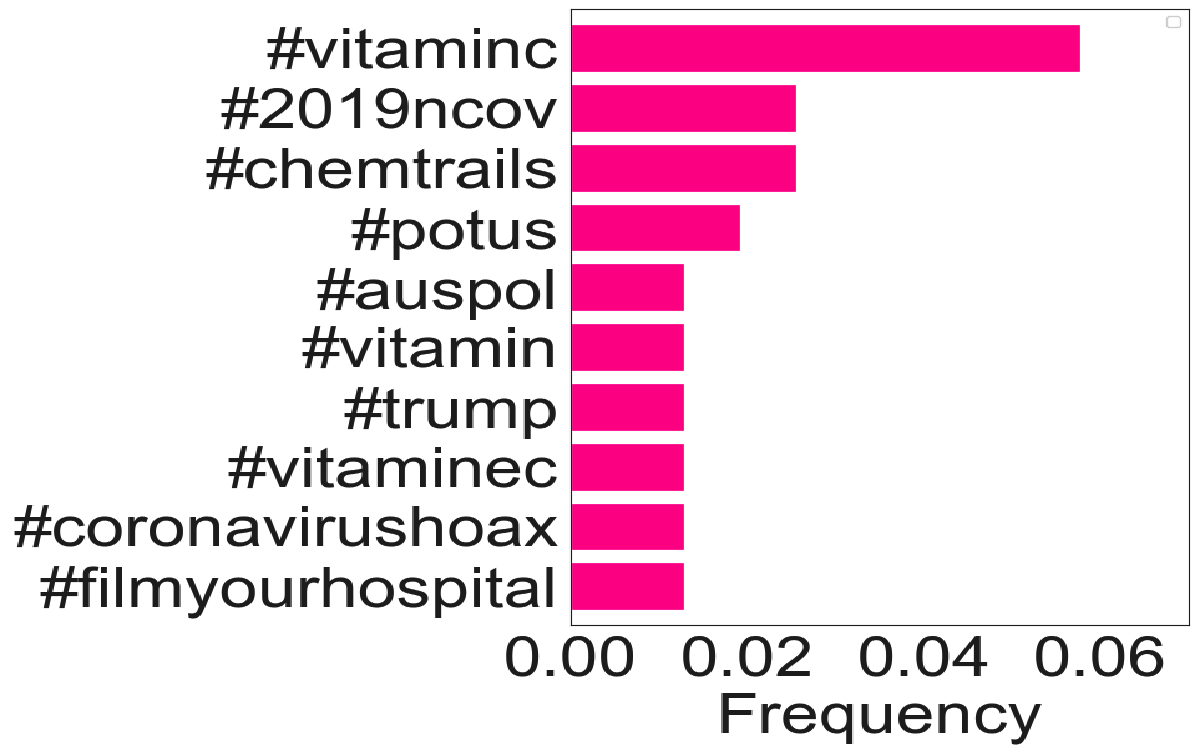}}
\subfigure[Highly severe]{
\includegraphics[width=0.4\columnwidth]{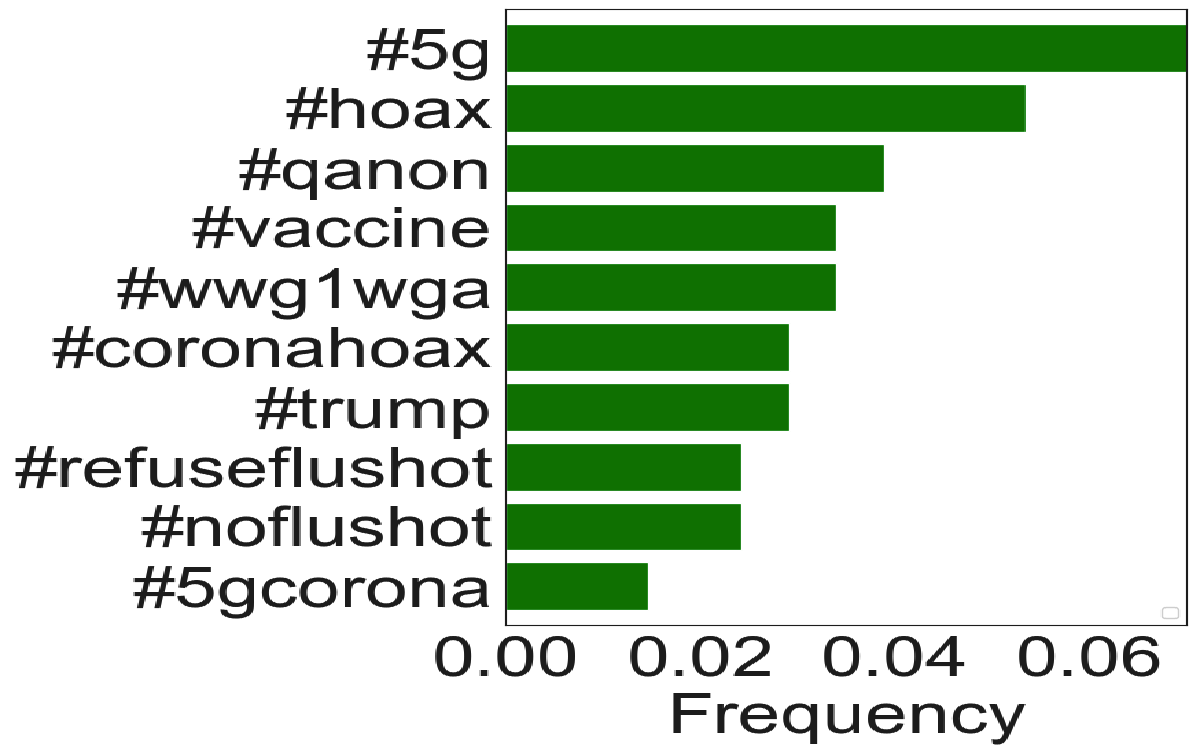}}
\subfigure[Refutes/Rebuts]{
\includegraphics[width=0.4\columnwidth]{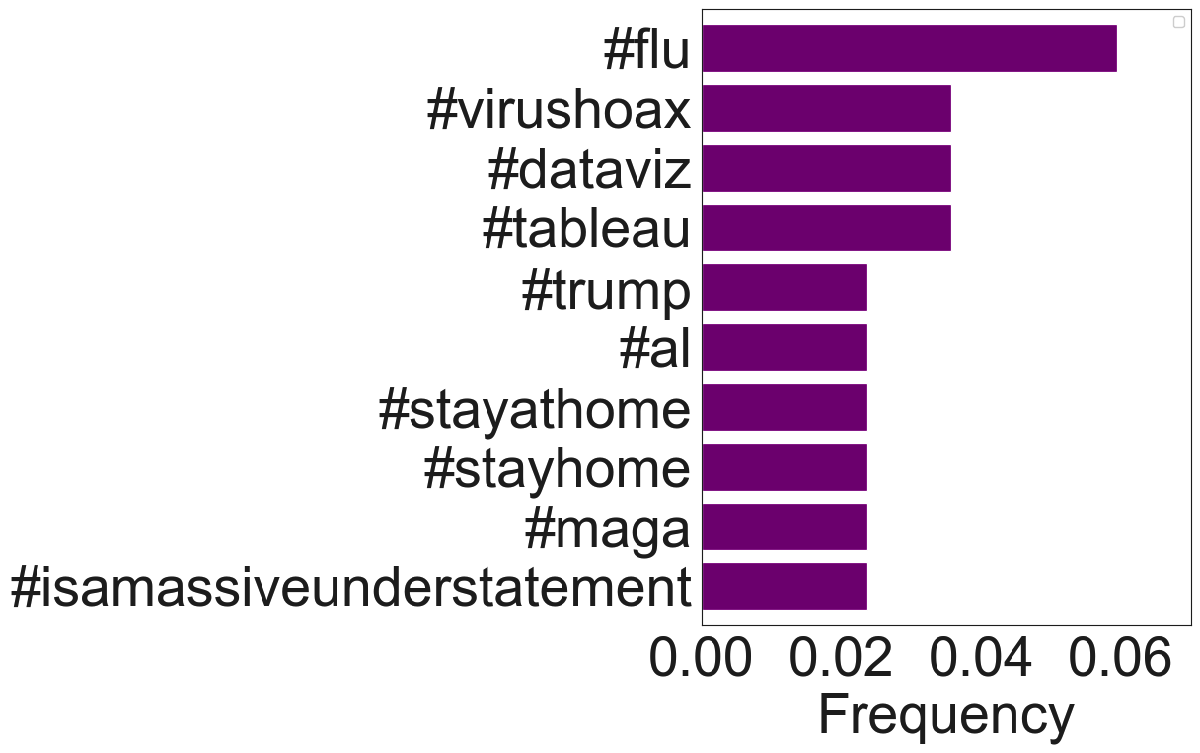}}
\subfigure[Other]{
\includegraphics[width=0.4\columnwidth]{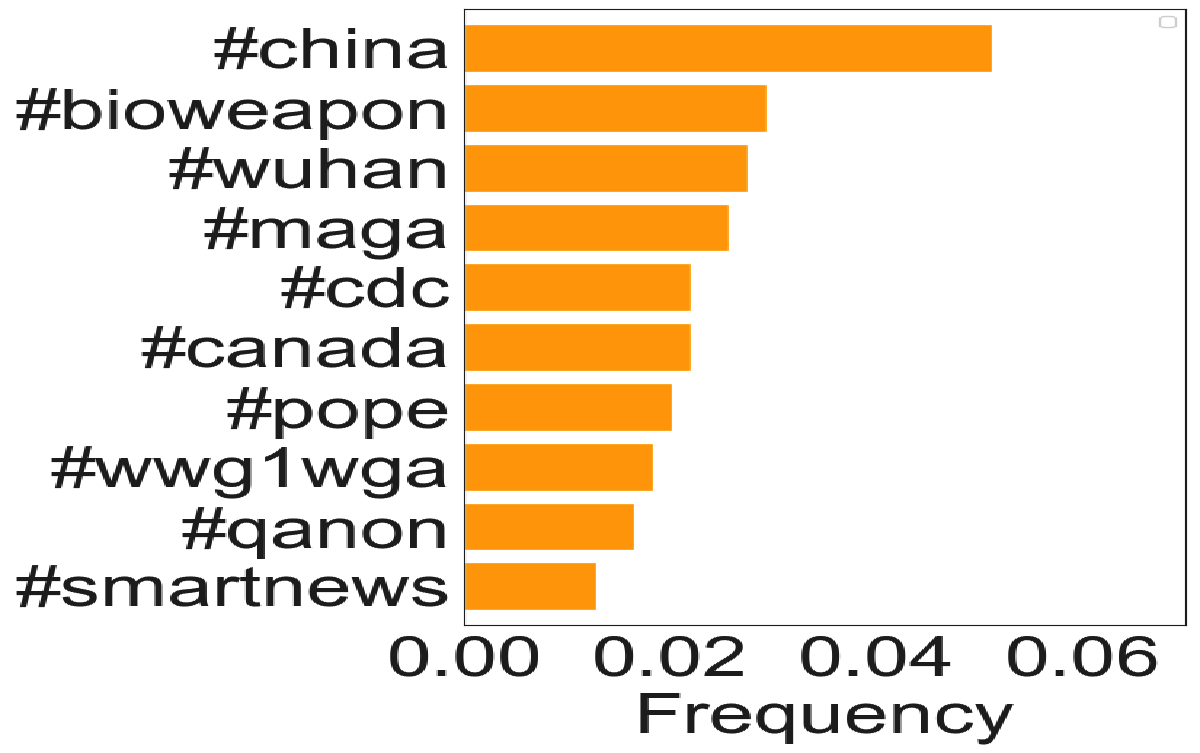}}
\subfigure[Real News/Claims]{
\includegraphics[width=0.4\columnwidth]{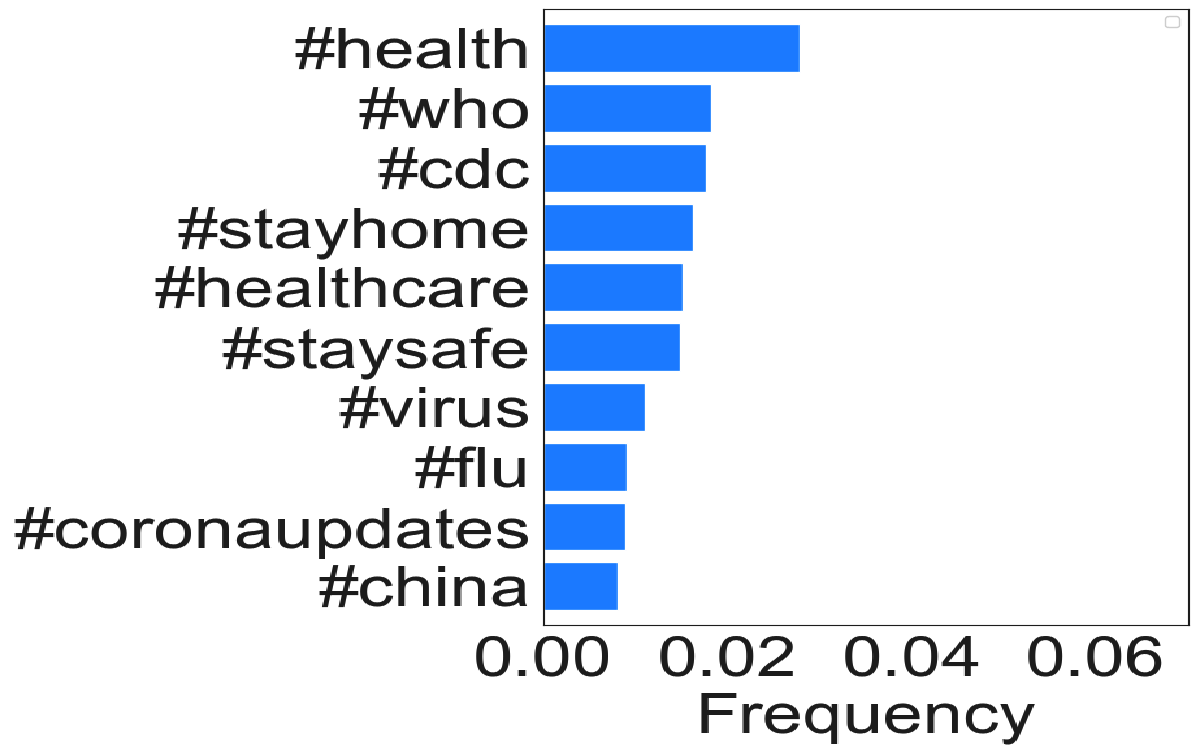}}
\centering
\caption{Most frequent hashtags per category.} 
\label{fig:hash}
\end{figure*}

In Figure \ref{fig:hash}, we analyze the top-10 frequent hashtags per category. We remove common hashtags such as ``\#covid\_19'', ``\#coronavirus'', \etc The length of each bar indicates how frequently the hashtag appears. We find that the ``Other'' category follows a similar pattern to Figure \ref{fig:freq}, in that the top hashtags are pertaining more to rumors and conspiracies, such as the ``\#pope'' tested positive for COVID-19, or that COVID-19 is a ``\#bioweapon''. This may be attributed to the fact that those susceptible to misinformation are less likely to think critically about news sources and thus tend to believe more false claims \cite{pennycook2020fighting}. Both severe categories focus on remedies, \eg ``\#vitaminc'' and vaccination. Interestingly, the  ``Refutes/Rebuts'' top hashtags contain terms associated with computation such as ``\#dataviz'', ``\#tableau'', as well as hashtags to promote social distancing ``\#stayhome''. These hashtags may be evidence of several infographics and data visualizations shared in social media, often used as arguments against misinformation. Finally, as per the aforementioned claim index (category claim for each tweet), we present the most common claims and news per category, and their corresponding frequency (Table \ref{tab:common}).

\section{Experiments}\label{sec:exp}
We perform experiments with several baselines and state-of-the-art models. We pre-process tweets to filter out reserved tokens, such as \textit{RT} or \textit{retweet}, urls and mentions. The same pre-processing is performed for tweet replies. Additionally, we split the data into $80\%$ training and $20\%$ testing, keeping the same splits across all models for a fair comparison. To avoid any data leakage, we remove duplicate tweets and retweets and stratify based on \textit{claim index}, \ie tweets discussing the same claim under the same label are kept either in training or test, but not in both. Our evaluation metrics are Precision (P), Recall (R) and F1 score (macro-averaged).

\noindent The algorithms we experiment with are:

\noindent \textbf{Random Forests} with bag-of-words (RF-TFIDF) or 100-dimensional pre-trained Glove embeddings  (RF-Glove) as text representation.

\noindent \textbf{Support Vectors} with bag-of-words (SVM-TFIDF) or 100-dimensional pre-trained Glove embeddings (SVM-Glove).

\noindent \textbf{Logistic Regression} with bag-of-words (LR-TFIDF) or 100-dimensional pre-trained Glove embeddings  (LR-Glove), same as for SVM and RF.

\noindent \textbf{Bi-directional LSTM}  model \cite{schuster1997bidirectional} with 100-dimensional pre-trained Glove embeddings as initial representation (BiLSTM). 

\noindent \textbf{Multichannel CNN} with multiple kernel sizes and 100-dimensional pre-trained Glove embeddings as initial representation, similar to \citet{kim2014convolutional} (CNN).

\begin{table*}[t!]
\centering
\resizebox{0.9\textwidth}{!}{%
  \begin{tabular}{p{0.65\linewidth}p{0.20\linewidth}p{0.1\linewidth}}
    \toprule
    \textbf{Claims/News} & \textbf{Category} &  \textbf{Frequency}\\
    \midrule
    COVID-19 testing (viral test procedures and information) &  Real News/Claims &  481 \\ \hline
    COVID-19 is more contagious than the flu &  Real News/Claims &  466\\ \hline
    Coronavirus Hoax &  Highly Severe &  338\\  \hline
    COVID-19 is just like the flu & Refutes/Rebuts &  287\\  \hline
    Lab-Made Coronavirus Triggers Debate &  Other &  152\\  \hline
    Michigan Governor Gretchen Whitmer Bans Buying US Flags $~~$ During Lockdown &  Other &  148\\  \hline
    Shanghai Government Officially Recommends Vitamin C &  Possibly severe &  102\\  \hline
    Flu Vaccine Increases Coronavirus Risk 36\% Says Military Study &  Highly severe &  99\\  \hline
    Vitamin C Protects Against Coronavirus & Possibly severe &  79\\  \hline
    Coronavirus [is not a] Hoax &  Refutes/Rebuts &  73\\ 
  \bottomrule
\end{tabular}
}
\caption{Most common claims with their corresponding frequency and ground-truth labels.}
\label{tab:common}
\end{table*}

\noindent \textbf{Task-specific BERT} fine-tuned on our downstream text classification task, initialized with general-purpose BERT embeddings \cite{devlin2018bert}. 

\noindent \textbf{Hierarchical Attention Networks} (HAN), with word-level and sentence-level attention mechanisms \cite{yang2016hierarchical}

\noindent \textbf{dEFEND}, a state-of-the-art fake news detection model that builds upon HAN by adding co-attention between two textual sources \cite{shu2019defend}; in our case either tweet replies or corresponding news content.
dEFEND allows us to evaluate whether incorporating additional sources of information with related content can improve predictive performance.

\subsection{Hyper-parameter Details}
Hyper-parameter search is performed with Tune \cite{liaw2018tune}, Async Successive Halving Scheduler \cite{li2018massively}, Adam \cite{kingma2014adam} and 50 mini-batch size for all models. We trained all the models, except BERT, for a total of $100$ epochs. We fine-tuned BERT for a total of $3$ epochs. The best performing hyper-parameters were: (1) For the bidirectional LSTM (BiLSTM), two hidden layers with $256$ neurons each and tanh activations, followed by the final classification layer. Dropout is set to  $0.5$. (2) The convolutional neural network (CNN) has $128$ filters of kernel size $\{3,4,5\}$, stride 1 and ReLU activations, followed by max-pooling, a dense layer of $256$ units, and the final classification layer. Dropout is set to  $0.5$.
(3) The hierarchical attention model (HAN) consists of a world-level GRU layer and a sentence-level GRU layer, each with $256$ neurons, tanh activations, followed by $0.7$ dropout and a final classification layer. dEFEND extends HAN with an additional GRU layer for the replies/news, with the same configuration as for the word and sentence GRU layers.
(4) BERT is fine-tuned by adding a $128$ linear layer and a final classification layer with ReLU activations and $0.7$ dropout. We train all deep learning models with two GeForce GTX $1080$ Nvidia GPUs on an $64$-bit Ubuntu $16.04.6$ machine with Intel(R) Xeon(R) CPU E5-2650@2.20GHz and $\sim 260$ GB RAM.

\subsection{Experimental Results}
In terms of F1 score, BERT performs better than all baselines (Table \ref{table_results}). Surprisingly, incorporating user engagement features or news content did not help\footnote{We additionally experiment with COVID-Twitter-BERT, a Transformer model pre-trained on 22.5M COVID-19 Twitter messages \cite{muller2020covid}, but we have observed lower performance than with general BERT. We leave further analysis to future work.}. One of the main challenges of the finer granularity in \texttt{Covid-HeRA} is that some categories might be substantially underrepresented. In other words, this task is a highly imbalanced problem. We evaluate most commonly used methods for mitigating class imbalance, including data augmentation for minority classes and custom loss functions\footnote{Presented in more detail in the appendix.}.  We find that replacing Cross-Entropy with Tversky loss \cite{salehi2017tversky} results in $6\%$ relative performance gains (Table \ref{table_results} - \textbf{BERT + Tversky}).
In addition, we experiment with word-level augmentations (oversampling), such as word insertions and substitutions based on pre-trained embeddings, and synonym replacement based on lexicons and thesauri (\eg WordNet) \cite{zhang2015character,wang2015s,fadaee2017data,kobayashi2018contextual}. We find that such augmentations produced the best performance (Table \ref{table_results} - \textbf{BERT + DA}) with $8\%$ relative performance gains.  

\begin{table}[t!]
\centering
\resizebox{\columnwidth}{!}{%
\begin{tabular}{p{2.9cm}*{3}{p{0.6cm}<{\centering}} ||*{3}{p{0.6cm}<{\centering}}}
\toprule
    \multirow{2}{*}{Method}
& \multicolumn{3}{c}{\textsc{\textbf{Multi-class}}} & \multicolumn{3}{c}{\textsc{\textbf{Binary}}} \\
\cmidrule(l){2-4} \cmidrule(l){5-7} 
& \textbf{P} & \textbf{R} & \textbf{F1} & \textbf{P} & \textbf{R} & \textbf{F1}  \\
\midrule
    RF-TFIDF  &  0.29 & 0.24 & 0.25 &  0.69 & 0.58 & 0.61 \\
    RF-Glove  &  0.26 & 0.20 & 0.20 &  0.65 & 0.50 & 0.49 \\
    SVM-TFIDF &  0.24 & 0.20 & 0.20 &  0.71 & 0.58 & 0.61 \\
    SVM-Glove &  0.33 & 0.23 & 0.25 &  0.69 & 0.58 & 0.61 \\
    LR-TFIDF &  0.28 & 0.22 & 0.23  &  0.68 & 0.58 & 0.63 \\
    LR-Glove &  0.31 & 0.23 & 0.25 &  0.71 & 0.57 & 0.60 \\ \midrule
    BiLSTM &  0.40 & 0.27  & 0.27 &  0.86 & 0.58 & 0.63 \\ 
    CNN    &  0.34 & 0.28 & 0.29 &  0.85 & 0.62 & 0.67 \\
    BERT &   0.34 & 0.32 & {\textbf{0.33}}  &   0.85 & 0.62 & 0.66 \\
    HAN &  0.20 & 0.20 & 0.20 &  0.65 & 0.67 & 0.67 \\
    dEFEND w. replies &  0.25 & 0.22 & 0.23 &  0.80 & 0.54 & 0.65 \\ 
    dEFEND w. news &  0.35 & 0.31 & 0.25 &  {\textbf{0.92}} & {\textbf{0.68}} & {\textbf{0.75}} \\
    \midrule
    BERT + DA &   {\textbf{0.55}} & {\textbf{0.51}} & {\textbf{0.41}} &  0.93 & 0.64 & 0.70\\
    BERT + Tversky &  0.40 & 0.48 & 0.39 &  0.68 & 0.83 & 0.72 \\
  \bottomrule
\end{tabular}
}
\caption{Performance of evaluated models on \texttt{Covid-HeRA}: proposed severity categorization (Left) and traditional misinformation (binary classification) task, \ie with posts labeled as misinformation or not (Right)}
\label{table_results}
\end{table}

Different from existing misinformation problem formulations, \texttt{Covid-HeRA} is more challenging. We perform the same experiments in a traditional binary misinformation detection setting, \ie predict the veracity of information. More specifically, we discard all refutation and rebuttal tweets and collapse all tweets labeled as misinformation in a common label irrespective of severity, \ie essentially backtracking to a real versus fake news traditional framework. We evaluate the same set of algorithms. Compared to the perceived severity classification task of \texttt{Covid-HeRA}, the traditional misinformation detection task produces much higher performance across all evaluation metrics (Table \ref{table_results}, right). These results leave much room for improvement, showcasing that this task is not yet fully addressed by current state-of-the-art models and highlighting the challenges of accurately distinguishing between general misinformation, harmful social media posts and refuted claims.

\subsection{Error Analysis and Discussion}
In order to identify potential avenues of improvement and limitations of the classifiers explored, we take a closer look at some instances (tweet messages) that are consistently misclassified across the set of evaluated deep learning classifiers, \ie instances that all neural models are not able to predict correctly. We also perform error analysis on the false negatives from the best performing classifier. We group false negatives based on the misinformation topic referenced in each message. 
For example, for the category ``Other'', which makes up the majority of the false negatives, test instances can be further categorized by the message topics, as shown in Table \ref{tab:er1}. In particular, in the \textit{Misclassified} column, we count the total number of messages that are incorrectly classified by all deep-learning models explored. Our error analysis shows that considering the different topics could potentially improve misinformation detection models on this task.

We also qualitatively examine the false positives of the best performing model (\ie BERT) and find only 11 false positives classified as misinformation, showing the classifier is conservative but precise in the classification. Future directions would include improving the recall, as the examples missed are usually controversial topics or messages which appear as misinformation but are actually opinions, such as the mean incubation period of the virus and the seriousness of the virus compared to the flu. There is a general trend of poor recall across the levels of severity, among all of the explored classifiers, which highlights the difficulty of predicting the severity of misinformation.

\section{Conclusion and Future Work}\label{sec:conc}
In this work, we present \textbf{Covid} \textbf{He}alth \textbf{R}isk \textbf{A}ssessment (\textbf{\texttt{Covid-HeRA}}), a novel task that aims to capture the perceived severity of health-related misinformation. We describe our data collection and conduct thorough data analysis and extensive experiments with baseline methods and state-of-the-art models. Our experimental results demonstrate the challenges that computational models face in capturing the true latent semantics in our finer-grained health misinformation dataset. We hope \texttt{Covid-HeRA} will significantly facilitate the research in modeling the perception of misinformation severity from the target audience and will spur the development of more advanced systems that can inform social media users on the respective dangers of following unreliable advice from inaccurate sources. 

\begin{table}[t!]
    \centering
    \resizebox{0.9\columnwidth}{!}{%
    \begin{tabular}{lcc} 
    \toprule
        Topic & Misclassified & Total \\ \midrule
        COVID-19 was artificially engineered & 56 & 72 \\
        Media COVID-19 scare & 25 & 33 \\
        False Death Toll Reports & 5 & 23 \\
        Trump Reelection & 8 & 9\\
        Surgeon General Projection Reports & 6 & 8 \\
    \bottomrule
    \end{tabular}}
    \caption{Topic themes in test data for the ``Other'' class. \textit{Misclassified} column: number of instances incorrectly classified by all neural classifiers. \textit{Total} column: total number of instances in test data. Both categorized by topic theme.}
    \label{tab:er1}
\end{table}
There are several new directions and opportunities. \texttt{Covid-HeRA} aims at a deeper understanding of reader views on misinformation severity and can have immediate practical impact, as the potential data user group includes computational linguists, sociologists, psychologists, misinformation researchers, \etc Our dataset captures how social media users, the main consumers of misinformation, view and recognize (harmful) fake news and claims. Such new directions that target whether readers understand the impact of misinformation can help reduce misinformation spread. 
 
In particular, the data collected includes a special class ``refutes/rebuts claim'' that can be used in conjunction with the replies and the news titles, to form a targeted tailor-made refutation of a severe false claim.  Since we release the tweet identifiers and code for downloading the raw tweets, all associated metadata can be acquired, \eg timestamp, likes, retweets, reply time, geolocation, \etc As such, \texttt{Covid-HeRA} can be useful for advancing propagation-based methods that incorporate metadata. While our work evaluates content-based methods for early detection, we hope future work will expand predictive modeling methods by incorporating metadata.

The temporal evolution of misinformation is another interesting application to our dataset, which can be accomplished through chronological ordering of tweets based on timestamps and trend forecasting, or by utilizing transfer learning and additional Twitter misinformation datasets on related health domains. Combined with additional resources, \texttt{Covid-HeRA} can be used for addressing research questions on how misinformation and perceived severity affects consumer trends during the pandemic, for example in regards to time-sensitive products such as masks, hand sanitizers, health supplements and even fake remedies.

To alleviate the need for large training sets, future research could focus on weakly-supervised, semi-supervised, and self-supervised algorithms. Few-shot models can also handle distribution shift and classes with fewer examples. Anomaly and outlier detection methods for text data are interesting  directions worthy of future exploration that could potentially improve the detection accuracy. As the presence of severe misinformation tweets is rare, our dataset is ideal for benchmarking text-based anomaly/outlier detection model architectures in realistic scenarios \cite{ruff2019self,manolache2021date}. Finally, the subtask of identifying rebuttal and refutation posts that present useful arguments against misinformation spread, is something we aim to further improve in the future, either by incorporating additional linguistic signals and auxiliary tasks, or by applying controversy detection algorithms \cite{lourentzou2015hotspots,timmermans2017computational}.

\section{Broader Impact}\label{sec:ethical}
This work aims at understanding how average social media users perceive the severity of misinformation. Our labels are produced by readers rather than medical experts, and hence are not suitable for traditional health analytics and misinformation fact-checking algorithms. Moreover, the detection of misinformation spreaders, as well as additional aspects in which a post might be dangerous, including but not limited to political implications, bias/racism towards specific groups, distrust in science, technological impact, \etc, go beyond our task but are certainly noteworthy for discussion and future research. 
All data crawled through the Twitter API fall under Twitter's privacy policy. Following the Twitter terms of service, we only distribute Twitter-provided identifiers to download the data, and refrain from publicly sharing any tweets or user information. We note that this paper describes a work of research, thus, any definitions and methods described here do not reflect how Twitter classifies misleading information or enforces its misinformation policies. We also note that users can choose to make their posts private, delete their accounts or remove tweets. Hence, as for all Twitter-related research, exact reconstruction of the dataset in later years may not be possible.
Misinformation spread on social media has been a long-standing problem with potential public health risks \cite{ghenai2018fake,geeng2020fake}. Considering the substantial increase in misinformation during the pandemic, and the lack of methods that aim at understanding the perceived severity of false claims and fake news, this work serves as a timely dataset on COVID-19 misinformation circulating on social media.

{\small \bibliography{main.bib}}

\section{Appendix}\label{sec:imbalance}

The dataset statistics show that there exists substantial imbalance in misinformation classes, compared to the real news/claims class (Table 2). This is a recurring challenge for misinformation tasks, as most tweets online are either irrelevant, opinions and discussions about health topics or news shared across the platform. 
Generally, there is an abundance of proposed methods in the literature that can handle class imbalance. The most common approaches involve either assigning weights to individual training examples, custom loss functions, re-sampling the data or generating synthetic examples to improve robustness \cite{chawla2002smote,xie2019unsupervised}. For a thorough review on class imbalance in machine learning, we refer the reader to literature reviews \cite{chawla2009data,he2009learning}. Here, we explore some of the aforementioned methods. Nevertheless, our goal is not to exhaustively experiment with all possible options.

In terms of data augmentation, we restrict our exploration to word-level augmentations, since, these methods produce good results for text classification tasks \cite{wei-zou-2019-eda,xie2019unsupervised}. In terms of custom loss functions, we experiment with a set of the most representative ones, including vanilla Cross-Entropy and weighted Cross-Entropy, in which examples are weighted disproportionately to the number of instances per class \cite{cui2019class}. 
We also experiment with the \textbf{Dice} and \textbf{Tversky} loss functions \cite{milletari2016v,salehi2017tversky} and their variants such as the \textbf{Soft Dice Loss} \cite{milletari2016v}, the \textbf{Generalized Dice Loss} \cite{crum2006generalized,sudre2017generalised} and the \textbf{Self-adjusting Dice Loss} \cite{li2019dice}.

\begin{table}[t!]
\centering
\resizebox{\columnwidth}{!}{%
\centering
\begin{tabular}{p{4.5cm}*{4}{p{0.6cm}<{\centering}}}
    \toprule
    & \textbf{P} & \textbf{R} & \textbf{F1} \\
    \midrule
    BERT + vanilla CE &  0.34 & 0.32 & 0.33  \\
    BERT + weighted CE & 0.38 & 0.34 & 0.35 \\
    BERT + Tversky & 0.40 & 0.48 & 0.39 \\ 
    BERT + Generalized Dice & 0.40 & 0.46 & 0.34 \\
    BERT + Self-Adjusting Dice & 0.44 & 0.45 & 0.36 \\
    BERT + Soft Dice & 0.38 & 0.40 & 0.37 \\  
    BERT + DA &  {\textbf{0.55}} & {\textbf{0.51}} & {\textbf{0.41}} \\
    BERT + DA + Tversky & 0.38 & 0.45 & 0.36\\  
    \bottomrule
\end{tabular}
}
  \caption{Comparison of methods that handle class imbalance on the \texttt{Covid-HeRA} data, with the best performing model, \ie BERT. Evaluation with Precision (P), Recall (R) and F1. }
    \label{tab:results_imbalance}
\end{table}

Regarding data-augmentation, the percentage of words in each sentence to be altered is set to $7\%$. The number of augmented samples per instance is $3$, \ie for each training instance, we create three additional synthetic examples by randomly sampling variations of word substitutions, insertions, deletions, \etc
For the Tversky loss we set $\alpha=0.6$ and $\beta=0.4$. For the Self-adjusting Dice loss, we set $\alpha=0.75$. In practice, a small smoothness factor is added to each loss. Here, we set this smoothness to $10^{-5}$.

In Table \ref{tab:results_imbalance}, we present results for all aforementioned methods, producing approximately $1-8\%$ relative performance gainS.
The vanilla Cross-Entropy cannot handle minority classes properly, mostly predicts all tweets as ``Real News/Claims'' and confuses ``Refutes/Rebuts'' with ``Highly Severe''. Both data augmentation (\textbf{BERT + DA}) and Tversky loss (\textbf{BERT + Tversky}) improve the prediction for some classes, \eg ``Other'' or ``Refutes/Rebuts'', however, some examples from the ``Possibly Severe'' and ``Highly Severe'' classes are misclassified as ``Real News/Claims''. This is possibly due to the broad range of similar topics, as described in our data analysis. Further research on handling such cases is required. We note that Tversky approximates our evaluation metric (F-measure) when $\alpha+\beta=1$ (as in our case). The combination of data augmentation and Tversky did not produce better results. In the future, we hope to introduce models that synergistically combine data augmentation and training losses.

\end{document}